\documentclass[letterpaper,journal]{IEEEtran}
\usepackage{amsmath,amsfonts}
\usepackage{algpseudocode}
\algrenewcommand\algorithmicrequire{\textbf{Input:}}
\algrenewcommand\algorithmicensure{\textbf{Output:}}
\algrenewcommand\alglinenumber[1]{\textbf{#1}}
\usepackage{algorithm}
\usepackage{array}
\usepackage[caption=false,font=normalsize,labelfont=sf,textfont=sf]{subfig}
\usepackage{textcomp}
\usepackage{stfloats}
\usepackage{url}
\usepackage{verbatim}
\usepackage{graphicx}
\usepackage{amssymb}
\usepackage{cite}
\usepackage{ragged2e}
\usepackage{threeparttable}
\usepackage{xcolor}
\usepackage{colortbl}
\usepackage{multirow}
\usepackage{booktabs}
\usepackage{microtype}
\usepackage[colorlinks=true,linkcolor=blue,citecolor=blue,urlcolor=blue,bookmarks=true]{hyperref}
\usepackage{xcolor}

\begin{document}

\title{GFSR: Geometric Fidelity and Spatial Refinement \\ for Reliable Lane Detection}


\author{
Tiancheng Wang\IEEEauthorrefmark{1}$^1$, 
Zhaolu Ding\IEEEauthorrefmark{1}$^1$, 
Richeng Xu$^1$, 
Tianhui Zheng$^1$, 
Hui Liu$^1$, 
Hanyu Xuan\IEEEauthorrefmark{2}$^{1,2,3}$,
Zhiliang Wu$^{4}$,
Guanghui Yue$^{5}$

\thanks{\IEEEauthorrefmark{1}Equal contribution.}%
\thanks{$^\dagger$Corresponding author (e-mail: \href{mailto:22176@ahu.edu.cn}{\textcolor{black}{22176@ahu.edu.cn}}).}
\thanks{
This work was supported  
in part by the National Natural Science Foundation of China (No.62302006 and No.12271002),
in part by the Natural Science Research Project of Anhui Educational Committee (No.2024AH040012).}
\thanks{
$^1$ the School of Big Data and Statistics, Anhui University, Hefei 230039, China.
$^2$ the School of Artificial Intelligence and Data Science, University of Science and Technology of China, Hefei 230026, China.
$^3$ Institute of Dataspace, Hefei Comprehensive National Science Center, Hefei 230088, China. 
$^4$ the College of Computing and Data Science, Nanyang Technological University, Singapore 639798, Singapore.
$^5$ the School of Biomedical Engineering, Shenzhen University Medical School, Shenzhen University, Shenzhen 518060, China.
}
}

\markboth{T.~WANG \MakeLowercase{\textit{et al.}}: GFSR: GEOMETRIC FIDELITY AND SPATIAL REFINEMENT FOR RELIABLE LANE DETECTION}{T.~WANG \MakeLowercase{\textit{et al.}}: GFSR: GEOMETRIC FIDELITY AND SPATIAL REFINEMENT FOR RELIABLE LANE DETECTION}

\maketitle

\begin{abstract}
Lane detection stands as a crucial perception task in autonomous driving and advanced driver assistance systems.
However, existing methods still degrade in complex real scenarios due to two major limitations.
First, classification confidence only characterizes the categorical existence of lane priors and has no strong correlation with geometric quality. 
If threshold filtering and NMS are conducted merely based on this confidence, 
the model tends to retain lane priors with high confidence while eliminating those with lower confidence but superior geometric representation.
Secondly, the regression modules in existing methods weaken correlations among sampling points, hindering fine-grained optimization of distant, high-curvature and complex-topology lanes and causing underfitting.
To address these issues, 
we propose \textit{Geometric Fidelity and Spatial Refinement} (GFSR), a framework consisting of \textit{LaneIoU-guided Confidence Calibration} (LCC) and \textit{Adaptive Gated Location Refinement} (AGLR).
Specifically,
LCC adopts LaneIoU as soft supervision to explicitly estimate the geometric fidelity of lane priors, which is further fused with classification confidence to construct the \textit{Collaborative Reliability Index} (CRI).
This index guides lane prior filtering, effectively retaining those with high classification confidence and favorable geometric quality.
Meanwhile, cooperating with regression heads in each refinement stage, AGLR predicts sampling point lateral offsets and adopts a gating mechanism to adaptively regulate correction magnitude, strengthen inter-point correlations and boost model adaptability as well as robustness toward complex lane scenarios.
Extensive experiments on CULane and CurveLanes demonstrate that our GFSR achieves state-of-the-art performance on CULane, with \mbox{F1$_{50}$} and \mbox{F1$_{75}$} scores of $\mathbf{81.46\%}$ and $\mathbf{65.01\%}$, and reaches $\mathbf{87.35\%}$ \mbox{F1$_{50}$} on CurveLanes.
\begingroup
\urlstyle{same}%
\hypersetup{urlcolor=magenta}%
The code will be available at {\itshape\url{https://github.com/71yvonne/gfsr-2d-lane-detection}}.%
\endgroup
\end{abstract}


\section{Introduction}
\label{Introduction}

\IEEEPARstart{L}{ane} detection aims to infer both semantic and positional information of each lane from the front-view image captured by an onboard monocular camera. 
As a critical component of the perception module in autonomous driving systems,
lane detection provides the essential basis for downstream decision-making and planning \cite{lach2024comprehensive}.
In practical deployments, particularly in advanced driver assistance systems (ADAS) \cite{luo2025deep}, 
lane detection serves  as an indispensable module supporting core functionalities such as lane keeping, lane departure warning, and lane‑level trajectory planning.

Early vision-based 2D lane detection methods 
\cite{xuan2017robust,aly2008real,wang2004lane,satzoda2010hierarchical} 
typically followed a fixed pipeline 
consisting of preprocessing, handcrafted feature extraction, filtering, edge detection, and final lane line fitting.
However, this reliance on manually crafted operators for feature extraction \cite{ma2002simultaneous} 
often results in limited accuracy and robustness in complex real-world scenes. 
To overcome these limitations, 
deep learning-based methods have gradually come to dominate the field, owing to their powerful representation capacity and superior performance \cite{bi2025lane}.

Depending on how lanes are represented,
existing deep learning-based methods can be divided into three categories: segmentation-based, parameter-based, and detection-based approaches \cite{luo2025deep}.
Segmentation-based methods \cite{neven2018towards,pan2018spatial,hou2019learning,zheng2021resa} 
treat lane regions as pixel-level masks, 
thereby formulating lane detection as a dense pixel-level classification problem.
In contrast, 
parameter-based methods \cite{tabelini2021polylanenet,liu2021end,feng2022rethinking,chen2023bsnet} 
model lanes via parametric equations,
regressing the curve coefficients that parameterize the lane geometry.
On the other hand,
detection-based methods
\cite{qin2020ultra,liu2021condlanenet,qin2022ultra,li2019line,Tabelini_2021_CVPR,zheng2022clrnet,honda2024clrernet,yang2024ldtr,chen2019pointlanenet,qu2021focus,wang2022keypoint} 
treat lanes as sparse structured detection targets, such as keypoints or anchors,
thereby transforming lane detection into a sparse localization problem.

Despite their respective advantages, 
each of these categories has inherent limitations.
For instance, 
segmentation-based methods 
incur high computational overhead and demand intricate post-processing. 
Meanwhile,
parameter-based methods are sensitive to coefficient perturbations and exhibit poor fitting reliability in complex scenarios. 
By comparison, 
detection-based methods avoid these drawbacks,
offering lower inference cost and greater adaptability to scene variations without enforcing fixed curve priors. 
Consequently, detection‑based methods achieve  a favorable  trade-off among structural flexibility, low-latency inference, computational efficiency, and robustness.

In detection-based methods,
lanes are typically detected by regressing or associating sparse, discrete targets \cite{he2024monocular}.
These targets fall into three types: keypoints denote start or end points\cite{chen2019pointlanenet,qu2021focus,wang2022keypoint}. Row anchors refer to lateral positions on predefined rows\cite{qin2020ultra,liu2021condlanenet,qin2022ultra}. Line anchors represent directional line segments \cite{li2019line,Tabelini_2021_CVPR,zheng2022clrnet,honda2024clrernet,yang2024ldtr}.

Among these, keypoint‑based methods inherently lack strong  structural constraints, as the keypoints provide only local information with limited global context.
On the other hand, 
row anchor‑based methods are constrained by per‑row single‑position prediction,
which limits their robustness under complex curvature and topology changes.
In contrast, 
line anchor‑based methods overcome these shortcomings by incorporating directional priors, thereby attaining robust global structure modeling.

The line anchor‑based methods 
reconstruct lane geometry by predicting point‑wise offsets at discrete sampling locations,
while jointly optimizing anchor classification and regression. 
Lane instances are subsequently obtained via confidence thresholding and Non‑Maximum Suppression (NMS).
This paradigm has been progressively advanced by a series of representative methods.
Line-CNN \cite{li2019line} generates  multi-directional line candidates  based on directional priors and offset regression. 
LaneATT \cite{Tabelini_2021_CVPR} introduces global context aggregation to alleviate missed detections and ambiguities caused by insufficient local cues. 
CLRNet \cite{zheng2022clrnet} further enhances localization accuracy and geometric consistency through cross-layer refinement and LineIoU constraints. 
CLRerNet \cite{honda2024clrernet} integrates LaneIoU into both target assignment and loss optimization, thereby enhancing the reliability of confidence scores.

\begin{figure*}[!t]
    \centering
    \includegraphics[width=\linewidth]{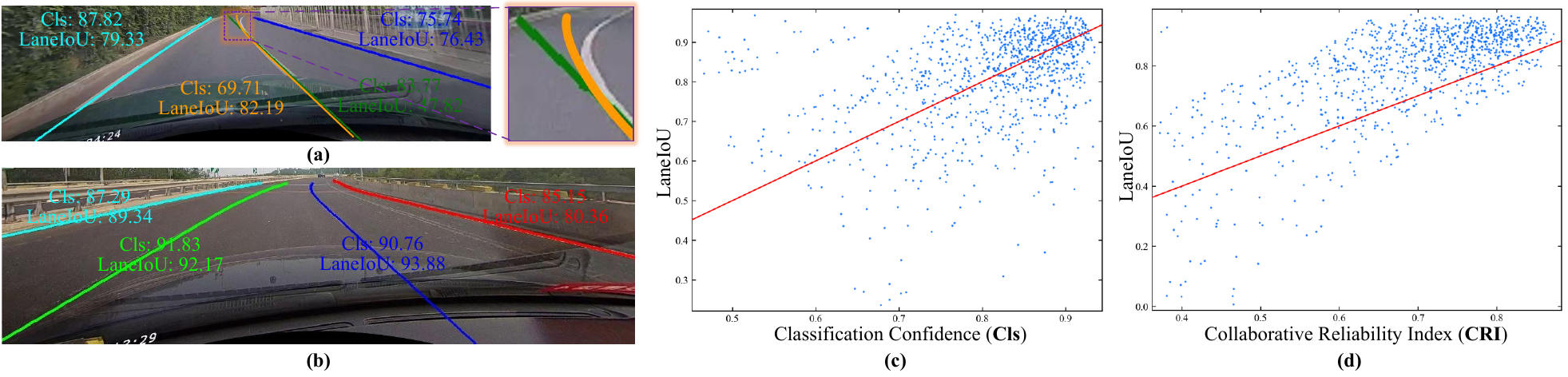}
    \caption{
    (\textbf{a}): An inconsistency between classification confidence (Cls) and geometric quality (LaneIoU) under high‑curvature scenarios. 
    Existing line anchor‑based approaches \cite{li2019line,Tabelini_2021_CVPR,zheng2022clrnet,honda2024clrernet} simply rely on Cls to rank and select lane candidates,  
    resulting in the \textcolor{green}{green} candidate preserved due to its higher Cls, despite its lower LaneIoU, while the \textcolor{orange}{orange} candidate is filtered out.
    (\textbf{b}): Lane candidates selected by the proposed method.
    (\textbf{c}): Correlation between LaneIoU and original Cls (Pearson coefficient: 0.40).
    (\textbf{d}): Correlation between LaneIoU and our CRI (Pearson coefficient: 0.69).}
    \label{fig:fig1}
\end{figure*}

Despite these advances, the above approaches \cite{li2019line,Tabelini_2021_CVPR,zheng2022clrnet,honda2024clrernet} 
simply rely on classification confidence derived from a classification head to rank and filter lane candidates in the post-processing stage.
When this confidence fails to correlate with actual geometric quality (i.e., LaneIoU),
candidates with high classification confidence but mediocre geometric quality tend to be preserved, whereas those with lower classification confidence yet superior geometric quality are incorrectly filtered out.
As illustrated in Fig.\ref{fig:fig1}(a), 
the orange candidate, despite its higher geometric quality, is suppressed due to lower confidence. 
Conversely, the green candidate, though  geometrically inferior, is preserved owing to its higher classification confidence.

The classification confidence is trained primarily to recognize the existence of lane candidates, rather than to assess their geometric quality.
Consequently, a discrepancy arises between the classification confidence and the actual geometric quality.
Fig.\ref{fig:fig1}(c) provides empirical evidence of this misalignment.
Although the overall trend between classification confidence and geometric quality exhibits a positive correlation,
the Pearson correlation coefficient is only moderate ($0.4$).
Moreover,
the candidate distribution is highly scattered. 
Notably,
certain candidates possess  high confidence yet show markedly low geometric quality. 
This inconsistency can induce false positives and false negatives during post-processing, thereby degrading overall detection performance.
Meanwhile, 
as highlighted in the magnified area of Fig.\ref{fig:fig1}(a),
we further observed that orange candidates, which should theoretically be retained, may still exhibit significant local fitting errors in
far-ahead and high-curvature lane  segments. 
This suggests  that merely  refining the global structural prior parameters is insufficient for fine-grained fitting of challenging lane segments, and may even introduce geometric errors.

To address the above issues, 
we propose \textit{Geometric Fidelity and Spatial Refinement} (GFSR), a framework for reliable lane detection. 
Our GFSR comprises a \textit{LaneIoU‑guided Confidence Calibration} (LCC) module and an \textit{Adaptive Gated Location Refinement} (AGLR) module,
which jointly calibrate classification confidence using LaneIoU supervision and 
adaptively refines the lane geometry with minimal computational overhead.
Specifically, 
the LCC module uses LaneIoU to supervise the learning of geometric fidelity and fuses this information with classification confidence. 
The fused confidence is defined as the \textit{Collaborative Reliability Index} (CRI).
This fusion mechanism fully takes into account the geometric reliability of lane candidates, 
thereby aligning the confidence used for thresholding, selection, and NMS more closely with the actual geometric quality.
The AGLR module performs gated, point-wise refinement of the offsets predicted at each sampling location.
This design amplifies the corrective modulation in regions of high curvature, occlusion, or topological ambiguity of lanes, while suppressing unnecessary adjustments in stable regions, thereby reducing the accumulation of local errors.

Owing to these designs,
as evidenced in Fig.\ref{fig:fig1}(d), 
the LCC module effectively calibrates classification confidence,
preserving more candidates that exhibit both high CRI and high LaneIoU,
which improves the Pearson correlation coefficient from $0.4$ to $0.69$.
Meanwhile,
as shown in Fig.\ref{fig:fig1}(b),
the AGLR module markedly improves lane geometry fitting in challenging scenarios involving high curvature.
Together, these two modules enable our GFSR to enhance the reliability of lane detection in complex scenes while maintaining detection accuracy and high inference speed.
The contributions of this work can be summarized as follows:
\begin{itemize}
    \item We propose geometric fidelity and spatial refinement, a lightweight framework designed to alleviate the inconsistency between classification confidence and geometric quality as well as the under-fitting of lanes in complex scenarios.
    \item We introduce the LaneIoU-guided confidence calibration module, which learns the geometric fidelity of lane candidates and fuses it with classification confidence to construct the collaborative reliability index,
    together with an adaptive gated location refinement module that refines point-wise lateral offsets.
    \item Extensive experiments demonstrate that our method achieves state-of-the-art performance on CULane, attaining \mbox{F1$_{50}$} and \mbox{F1$_{75}$} scores of $81.46\%$ and $65.01\%$, respectively. On CurveLanes, it achieves an \mbox{F1$_{50}$} score of $87.35\%$, while maintaining computational efficiency and real-time inference speed.
\end{itemize}

\section{Related Works}
According to the representation of lanes,
i.e., as pixel‑level masks, parametric equations, or structured formulations such as keypoints or anchors,
existing deep learning-based lane detection methods can be categorized  into  segmentation-based, parameter-based, and detection-based approaches \cite{luo2025deep}.

\subsection{Segmentation-based Methods}

Segmentation-based methods typically formulate lane detection
as a pixel-wise semantic classification problem, 
where each pixel is labeled to distinguish background from lane.
LaneNet \cite{neven2018towards} predicts per‑pixel lane segmentation probabilities and embedding vectors, which are then used to cluster pixels into a distinct lane instance.
SCNN \cite{pan2018spatial} enhances global interaction through layer-wise connections of features, enabling row-wise and column-wise information propagation.

However, these methods \cite{neven2018towards,pan2018spatial} incur substantial computational overhead, motivating subsequent works to focus on improving detection efficiency.
For example, SAD \cite{hou2019learning} employs an adaptive distillation mechanism to aggregate contextual information, thereby enabling the use of lightweight networks while maintaining competitive performance.
RESA\cite{zheng2021resa} leverages lane priors to capture row-wise and column-wise spatial relationships via a parallel shifting mechanism,
thus improving the efficiency of global information aggregation.

Although segmentation-based methods can achieve pixel-level lane localization via dense masks, 
they suffer from several intrinsic limitations.
These methods adopt dense prediction paradigms that prioritize local contextual hints, yet they lack the capability to efficiently characterize the overall geometric layout of lane structures.
Moreover, instance decomposition often requires clustering or heuristic post‑processing, which adds procedural complexity \cite{zhang2025asymmetric}.

\subsection{Parameter-based Methods}

Parameter‑based methods represent lanes using appropriate mathematical expressions, typically curve equations in image space. They regress curve parameters and then fit lane lines accordingly.
For instance,
PolyLaneNet \cite{tabelini2021polylanenet} detects each lane by regressing deep polynomial coefficients, the domain of the output polynomial, and confidence scores. 
LSTR \cite{liu2021end} further introduces Transformer-based context modeling and geometric priors to capture long-range dependencies in complex scenes.

Feng et al. \cite{feng2022rethinking} argue that abstract polynomial coefficients are difficult to learn. 
For this purpose, they introduce third-order B\'ezier curves to represent lanes, where the network directly predicts four B\'ezier control points to determine lane positions without any post-processing steps. 
However, this approach suffers from accuracy degradation because different parts of the curve may compete with each other.
To overcome this limitation, BSNet\cite{chen2023bsnet} adopts  B-spline curves for lane prediction, which can capture both global and local shape characteristics while enabling efficient parameter optimization.

Compared to segmentation-based lane detection methods \cite{neven2018towards,pan2018spatial,hou2019learning,zheng2021resa}, 
parameter-based methods \cite{tabelini2021polylanenet,liu2021end,feng2022rethinking,chen2023bsnet} are often more end‑to‑end and generally rely less on heavy post‑processing \cite{zhang2025asymmetric}. 
However, they remain sensitive to parameter regression errors, where slight inaccuracies in the estimated parameters can be amplified along the curve, so the reconstructed polyline may deviate substantially from the intended geometry.

\subsection{Detection-based Methods}
Detection-based methods typically represent lanes as learnable, structured formulations, such as keypoints or anchors.
According to the type of detection target,
these methods can generally be categorized into three types \cite{he2024monocular}:
keypoint‑based methods with sparse points, row anchor‑based methods with lateral positions on predefined rows, and line anchor‑based methods with directional line segments.

\subsubsection{Keypoint-based Methods}
Keypoint-based methods treat lane detection as a joint task of keypoint localization and association, where lanes are represented solely by a set of structural keypoints.
For example,
PointLaneNet \cite{chen2019pointlanenet} directly predicts discrete keypoints along each lane and their lane‑wise categories. Subsequently, NMS is applied to remove redundant predictions, and the remaining keypoints are connected to fit lane geometries.
FOLOLane \cite{qu2021focus} further improves inference efficiency by combining high‑resolution heatmap prediction with a faster decoding strategy.
However, point‑by‑point association suffers from low efficiency, a lack of global context, and a tendency to accumulate errors. 
To address these limitations, GANet \cite{wang2022keypoint} adopts a global association strategy that regresses the offset from each keypoint to its lane start point, thereby enabling parallel keypoint grouping.

These keypoint-based lane detection methods \cite{chen2019pointlanenet,qu2021focus,wang2022keypoint} remain dependent on association quality and post‑processing. 
Under challenging conditions such as occlusion, worn lanes, or densely adjacent lanes,
missing or mismatched keypoints can propagate into lane‑level errors, thereby degrading recall and topological consistency.

\subsubsection{Row Anchor-based Methods}
Row anchor‑based methods represent lanes by their lateral grid position on each predefined row. They formulate lane detection as row‑wise classification, predicting the most likely lane grid for each row.
For instance,
Yoo et al. \cite{yoo2020end} compress horizontal spatial information into row‑wise features via horizontal reduction modules and then predict the lateral grid position of the lane.
UFLD \cite{qin2020ultra} further introduces a structural loss to enforce shape consistency, thereby enabling explicit use of lane information.
CondLaneNet \cite{liu2021condlanenet} adopts a top‑to‑down approach that first detects lanes instances via start point prediction, then uses conditional convolution to dynamically generate instance‑specific kernels for subsequent row‑wise shape prediction.
Nevertheless, these methods \cite{yoo2020end,qin2020ultra,liu2021condlanenet} are not suitable for handling curved or nearly horizontal lanes, as a single row may intersect multiple areas of the same lane.
To address this issue, UFLDv2 \cite{qin2022ultra} adopts mixed anchor points, using row anchors for the self-lane and column anchors for the opposite lane, which reduces amplified localization errors.

These row anchor‑based lane detection methods \cite{yoo2020end,qin2020ultra,liu2021condlanenet,qin2022ultra} can strike a favorable balance between accuracy and latency.
However, they struggle to model cross‑row continuity under severe occlusion or shadowing, which may result in lane discontinuities or missed detections.

\subsubsection{Line Anchor-based Methods}
Similar to row anchor-based methods \cite{yoo2020end,qin2020ultra,liu2021condlanenet,qin2022ultra}, line anchor-based lane detection methods introduce directional features such as angles to represent lanes as straight segments and
minimize the deviation between line anchors and ground‑truth lane positions. 
For instance,
Line-CNN \cite{li2019line} predefines multi‑angle candidate rays at the bottom, left, and right boundaries of the image and directly regresses discrete point sets of lanes in an end‑to‑end manner.
However, this method requires a sufficient number of output channels to predict parameters, such as the length and offset of all anchors. For this purpose, LaneATT \cite{Tabelini_2021_CVPR} introduces an anchor‑based attention mechanism to aggregate  global context, thereby reducing error accumulation caused by local information  and improving detection performance on blurry lanes or occluded lines.

To further enhance lane detection accuracy, 
CLRNet \cite{zheng2022clrnet} adopts a cross‑layer refinement architecture that first detects lanes using high‑level semantic features and then refines them with low‑level details. It also  proposes a LineIoU loss that regresses the lane line as a whole unit rather than individual points, improving overall detection consistency.
Built upon CLRNet,
CLRerNet \cite{honda2024clrernet} integrates LaneIoU into both the target assignment and the loss function, thereby enhancing the reliability of confidence scores.

These line anchor-based methods \cite{li2019line,Tabelini_2021_CVPR,zheng2022clrnet,honda2024clrernet} select candidates that align well with slender lane structures, effectively avoiding the single lateral position per row assumption inherent to row anchor-based approaches. 
Nevertheless, they still suffer from a persistent misalignment between classification confidence and geometric quality, along with insufficient point-level refinement capabilities for highly curved or complex lane geometries.

\section{Proposed Method}

\begin{figure*}[t]
    \centering
    \includegraphics[width=1.0\linewidth]{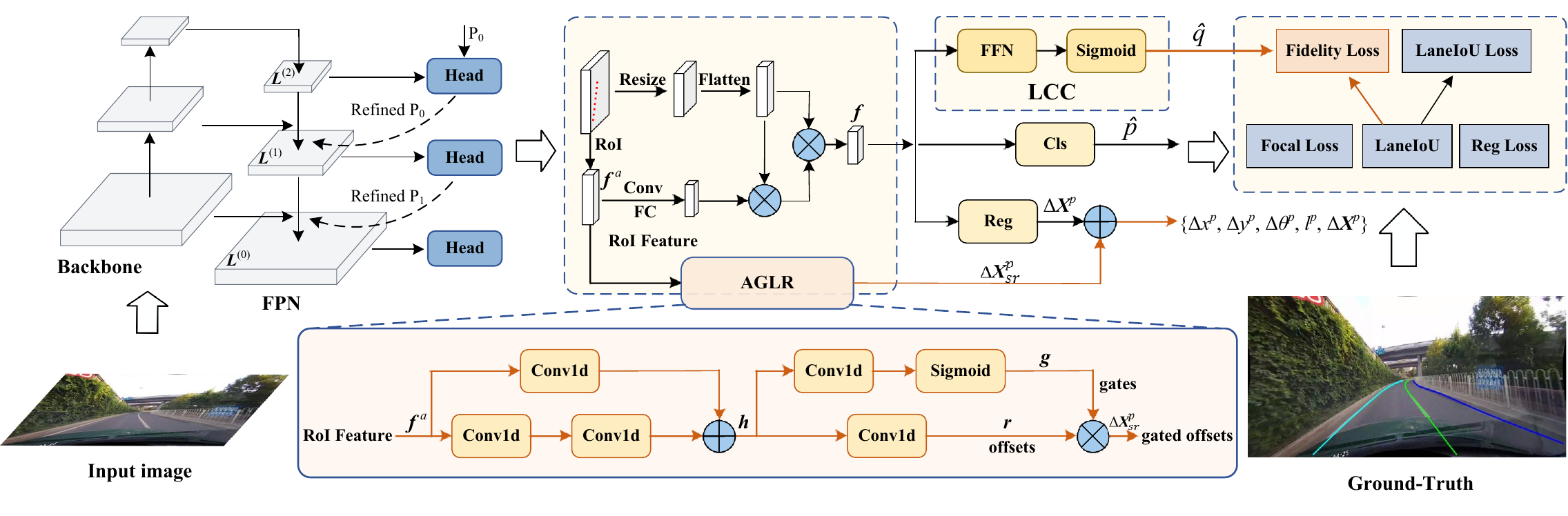}
    \caption{Overall framework of the proposed Geometric Fidelity and Spatial Refinement (GFSR), which comprises a LaneIoU‑guided Confidence Calibration (LCC) module and an Adaptive Gated Location Refinement (AGLR) module. The LCC predicts geometric fidelity and calibrates classification confidence with LaneIoU supervision, while the AGLR performs adaptive gated point-wise refinement of lane geometry, ensuring reliable lane detection with minimal computational overhead.}
    \label{fig:lfr_overview}
\end{figure*}

The overall architecture of our GFSR is illustrated in Fig.\ref{fig:lfr_overview}.
As shown in this figure, our GFSR follows the Cross Layer Refinement lane detection paradigm of CLRNet\cite{zheng2022clrnet}, which consists of a backbone, a Feature Pyramid Network
(FPN)\cite{Lin_2017_CVPR}, a RoIGather\cite{zheng2022clrnet} to extract lane RoI features and a cascaded cross-layer refinement detection head\cite{zheng2022clrnet}.
In the following,
we first introduce the representation of lanes, the refinement and assignment of lane priors,
and then present the designed LCC and AGLR in our GFSR.
Finally, we present all loss functions.

\subsection{Formulation and Overview}
\label{sec:A}

\subsubsection{Lane Representation}
In standard object detection\cite{murat2025comprehensive}, 
objects are typically represented by rectangular bounding boxes.
This representation is suboptimal for lane detection,
as lane lines are thin, elongated structures that exhibit strong geometric regularity.
To better capture these properties, 
we follow prior works \cite{li2019line,Tabelini_2021_CVPR,zheng2022clrnet,honda2024clrernet} and adopt a set of equally spaced 2D points as the lane representations,
where predefined parametric lane priors serve as structural templates to guide the network's predictions.

Given a front-view image $ I \in \mathbb{R}^{H_v\times W_v\times 3}$, 
we sample each lane at $N$ uniformly spaced vertical coordinate 
$y_i=\frac{H_v}{N-1}i$.
Each lane is then compactly described  by a parameter set $G = \{(x^g,y^g), \theta^g, l^g, \boldsymbol{X}^g\}$,
which consists of the entry point $(x^g,y^g)$,
the direction angle $\theta^g$, 
the number of valid points $l^g$,
and a series of horizontal positions $\boldsymbol{X}^{g}=\{x_i^g\}_{i=0}^{N-1}$ along these sampled rows.
We define lane priors as follows.
Each prior is parameterized by a starting point $(x^p,y^p)$
and an orientation angle $\theta^p$,
which are optimized end-to-end jointly with the detector. 
The horizontal coordinates of the prior are denoted as 
$\boldsymbol{X}^{p} = \{x_{i}^{p}\}_{i=0}^{N-1}$.

\subsubsection{Lane Prior Refinement}
As depicted in Fig.\ref{fig:lfr_overview},
the input is first processed by the backbone and FPN to produce multi-scale features $\{\boldsymbol{L}^{(s)}\}_{s=0}^2$. 
The refinement proceeds in a top-down manner.
At refinement stage $s$, 
the detection head performs RoI pooling along each lane prior to extract an anchor-level feature $\boldsymbol{f}^a$. 
This feature is then enhanced by convolutional layers to capture fine‑grained spatial structures, 
and subsequently  projected through fully connected layers for semantic compression. 
In parallel, the feature $\boldsymbol{L}^{(s)}$ is resized and flattened into a global vector.
Ultimately, 
the anchor-wise and global features are aggregated  through an attention mechanism, yielding the final  embedding $\boldsymbol{f}$.

For each lane prior, the regression head predicts the geometric offsets $(\Delta x^p, \Delta y^p)$, $\Delta\theta^p$, $\Delta{\boldsymbol{X}}^p$, together with an effective sampling length $l^p$.
The prior parameters are then updated by
\begin{equation}
\begin{aligned}
    (x^p, y^p) &\leftarrow (x^p + \Delta x^p,\; y^p + \Delta y^p), \\
    \theta^p &\leftarrow \theta^p + \Delta\theta^p, \\
    \boldsymbol{X}^p &\leftarrow \boldsymbol{X}^p + \Delta \boldsymbol{X}^p.
\label{eq:parameter update}
\end{aligned}
\end{equation}

\subsubsection{Lane Prior Assignment}
After the lane prior parameters are updated,
we compute the Lane Intersection-over-Union (LaneIoU) between each refined lane prior $P$ 
and each ground-truth (GT) lane $G$.
The cost matrix $\mathbf{C}$ is constructed using the normalized LaneIoU scores, with entries 
\begin{equation}
c_{j,k} = - LaneIoU(P_j, G_k)
+ \lambda F_{class}(P_j, G_k),
\end{equation}
where $P_j$ and $G_k$ represent the $j^{th}$ prior and the GT lane $k$, respectively. $F_{class}$ computes the classification cost between $P_j$ and $G_k$.

For each GT lane,
we then dynamically assign an adaptive number of lane priors by selecting the $K$ candidates with the lowest costs in the $k^{th}$ column of $\mathbf{C}$.
The matched lane priors constitute the positive set $\mathcal{M}$, while non-matching priors form the negative set $\mathcal{N}$. 
The classification label $Y_j$ for prior $P_j$ is then defined as $Y_j = 1$ if $P_j \in \mathcal{M}$, and $Y_j = 0$ otherwise.
For each positive prior, we also record the index of its matched GT lane, 
which provides the supervision for both regression and geometry-related losses.

\subsection{LaneIoU‑guided Confidence Calibration}
\label{lcc}

At refinement stage $s$, 
the classification head applies Feed-Forward Networks (FFN) followed by softmax to map the final embedding $\boldsymbol{f}$ to a classification confidence for the $j^{th}$ lane prior.
\begin{equation}
    \boldsymbol{\hat{p}}_j = \mathrm{softmax}\big(\mathrm{FFN}(\boldsymbol{f})\big) = [1-\hat{p}_j, \hat{p}_j],
    \label{Eq3}
\end{equation}
where $\hat{p}_j$ denotes the scalar classification confidence.
During inference,
${\hat{p}}_j$ is primarily utilized for post-processing.
Specifically, given a predefined confidence threshold $\tau$,
lane priors with ${\hat{p}}_j \geq \tau$
are retained as decoding candidates.
Geometric decoding then converts the predicted coordinates on the sampled rows into an ordered polyline in the image plane.
Among these candidates, 
${\hat{p}}_j$ serves as the ranking score for NMS and Top-$K$ selection, 
ensuring that overlapping hypotheses preferentially retain instances with higher confidence.

\subsubsection{Motivation}
During training,
the classification head 
is optimized with binary cross-entropy supervision.
Its objective is limited to distinguishing whether a given lane prior corresponds to a lane, without explicitly constraining the geometric discrepancy between the candidate prediction and the ground-truth.
Consequently,
the resulting classification confidence  ${\hat{p}}_j$ primarily indicates lane existence, rather than offering a comprehensive assessment of geometric quality.

From a decision-theoretic perspective \cite{ferguson2014mathematical}, 
relying solely on classification confidence for ranking and filtering is suboptimal.
The ideal detection score $W_j$ should capture both lane presence $Y_j$ and geometric quality $q_j \in [0,1]$ (e.g., LaneIoU):
\begin{equation}
W_j = Y_j\cdot q_j.
\label{eq:detection_criterion}
\end{equation}
Accordingly, the optimal objective should maximize the conditional expected score:
\begin{equation}
\mathbb{E}[W_j\mid \boldsymbol{f}]
=
P(Y_j=1\mid \boldsymbol{f})\cdot
\mathbb{E}[q_j\mid Y_j=1,\boldsymbol{f}].
\label{eq:E}
\end{equation}
Specifically, $P(Y_j=1\mid \boldsymbol{f})$ reflects the estimated probability that the lane prior corresponds to a lane, whereas $\mathbb{E}[q_j\mid Y_j=1,\boldsymbol{f}]$ captures the expected geometric quality given that the prior is a lane, as formalized by conditioning on $Y_j=1$.

In prior works \cite{li2019line,Tabelini_2021_CVPR,zheng2022clrnet,honda2024clrernet},
only the classification confidence ${\hat{p}}_j$ from Eq.\ref{Eq3} is employed  for ranking and filtering.
This practice implicitly treats the $\mathbb{E}[q_j\mid Y_j=1,\boldsymbol{f}]$ in Eq.\ref{eq:E} as a constant across all lane priors, i.e. it assumes all priors possess identical geometric quality. 
However, this assumption is frequently violated in real-world challenging scenarios, including occluded lanes, low-light conditions, and complex multi-lane intersections.

\subsubsection{Details}

For this purpose, we propose a LaneIoU-guided
Confidence Calibration (LCC) module. 
Our LCC leverages lane-level geometric overlap (i.e., LaneIoU) as a soft supervision to explicitly estimate the geometric fidelity of each lane prior.  
During inference, we fuse geometric fidelity and classification confidence into a Collaborative Reliability Index (CRI) for threshold filtering and NMS ranking.

As illustrated in Fig.\ref{fig:lfr_overview}, the proposed LCC is integrated into the detection head in parallel with the classification and regression heads. 
At each refinement stage, 
we introduce a lightweight FFN to map the final embedding $\boldsymbol{f}$ 
to a scalar $\hat{q}_j \in [0,1]$ via a sigmoid activation function:
\begin{equation}
    \hat{q}_j = \mathrm{sigmoid}\!\left(\mathrm{FFN}(\boldsymbol{f})\right).
    \label{Eq6}
\end{equation}
The resulting $\hat{q}_j$ represents the predicted geometric fidelity of the $j^{th}$ lane prior.

In lane detection \cite{li2019line,Tabelini_2021_CVPR,zheng2022clrnet,honda2024clrernet}, the geometric localization accuracy of predicted lanes is typically measured by LaneIoU between prediction and GT lane. 
Accordingly,
we employ LaneIoU as a soft supervision label for geometric fidelity.
Specifically,
for a lane prior $P_j$ matched to GT lane $G_{k}$,
its soft label $q_j$ is defined as
\begin{equation}
q_j = 
\begin{cases}
\mathrm{LaneIoU}(P_j, G_k), & \text{if}~~ P_j \in \mathcal{M},\\
0, & \text{if}~~ P_j \in \mathcal{N}.
\end{cases}
\label{Eq7}
\end{equation}
We adopt the Binary Cross-Entropy (BCE) loss to penalize discrepancies between the predicted geometric fidelity $\hat{q}_j$ and the soft label $q_j$.
The stage-wise fidelity loss is defined as
\begin{equation}
    L_{fid}^{(s)} = \frac{1}{|\mathcal{M}|}\sum_{j\in\mathcal{M}}\mathrm{BCE}\!\left(\hat{q}_j,q_j\right)+\frac{1}{|\mathcal{N}|}\sum_{j\in\mathcal{N}}\mathrm{BCE}\!\left(\hat{q}_j,q_j\right).
\label{eq:fidelity_loss}
\end{equation}
The overall  fidelity loss $L_{fid}$ is obtained by averaging the stage-wise $L_{fid}^{(s)}$ over all refinement stages, and is then incorporated into the overall multi‑task training objective.

During inference, 
we compute the CRI for each lane prior $P_j$ by fusing its classification confidence $\hat{p}_j^*$ defined in Eq.\ref{Eq3} and geometric fidelity $\hat{q}_j^*$ defined in Eq.\ref{Eq6} from the final refinement stage:
\begin{equation}
\mathrm{CRI}_j = \hat{p}_j^{*} \cdot \left(\beta_0 + \beta_1 \hat{q}_j^{*}\right),
\label{eq:fidelity_score_fusion}
\end{equation}
where $\beta_0$ and $\beta_1$ are non-negative calibration coefficients.
As indicated in Eq.\ref{eq:fidelity_score_fusion},
this affine mapping with respect to $\hat{q}_j^{*}$ allows  $\beta_0$ to  scale the direct contribution of the classification confidence, while $\beta_1$ modulates the strength of the geometry‑aware correction.
In post‑processing, CRI replaces the original classification confidence for both threshold filtering and NMS ranking. 
By jointly capturing lane existence (via $\hat{p}_j^*$) and geometric fidelity (via $\hat{q}_j^*$), CRI closely approximates the ideal detection criterion $W_j = Y_j \cdot q_j$ defined in Eq.\ref{eq:detection_criterion}, thereby enabling more reliable candidate selection.

\subsection{Adaptive Gated Location Refinement}

\subsubsection{Motivation}
The regression head feeds the final embedding $\boldsymbol{f}$ into a feed-forward network (FFN) to predict the horizontal offset $\Delta{\boldsymbol{X}}^{p}$ of each lane prior, and updates the horizontal coordinate $\boldsymbol{X}^{p}$ following Eq.\ref{eq:parameter update}.
Nevertheless, this design still depends on a single global final embedding $\boldsymbol{f}$ to uniformly predict the geometric parameters of the entire prior. 
The regression head struggles to explicitly leverage point-level information and model point-wise dependencies, 
which degrades geometric prediction accuracy and easily causes local fitting errors, 
especially in scenarios with complex road layouts.

In the post-processing stage, the model needs to maximize the expected detection score, 
where the $\mathbb{E}[q_j\mid Y_j=1,\boldsymbol{f}]$ defined in Eq.\ref{eq:E}
characterizes the expected detection quality. 
However, local underfitting resulting from the lack of point-level modeling reduces 
$\mathbb{E}[q_j\mid Y_j=1,\boldsymbol{f}]$,
which further weakens the reliability of detection scores and deteriorates the overall model performance.

\subsubsection{Details}
To mitigate this limitation, we introduce an Adaptive Gated
Location Refinement (AGLR) module that operates alongside the original regression head at each refinement stage.
It performs finer-grained point‑wise offset prediction and simultaneously estimates a confidence gate for each sampling point. 
By applying this gating mechanism, AGLR enhances local geometric continuity while preserving the reliability of the global lane structure.

\begin{figure}[t]
    \centering
    \includegraphics[width=1.0\linewidth]{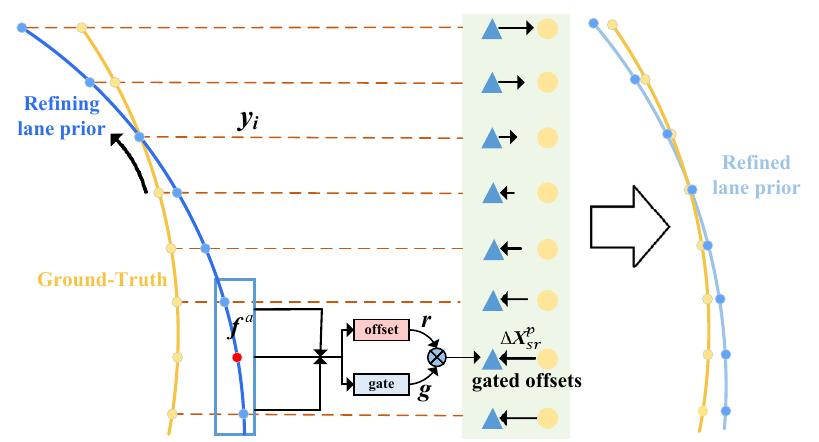}
    \caption{Schematic diagram of the AGLR module for refining lane prior: after RoI pooling, the anchor-level feature $\boldsymbol{f}^a$ is decoded to output offset and confidence gate.
    The final gated offset is obtained by element-wise multiplication of offset and gate, enabling adaptive correction of lane priors.
    In the gated offset representation, the arrow direction indicates the correction direction, and the arrow length represents the correction magnitude.
}
\label{fig:lanerefine_spatial}
\end{figure}

Fig.\ref{fig:lanerefine_spatial} illustrates the correction process of the proposed AGLR module.
We adopt a lightweight 1D convolutional residual structure to process the RoI-pooled anchor-level feature $\boldsymbol{f}^a$.
The main path consists of two $3\times1$ 1D convolutions that capture context along the sampling dimension, while a $1\times1$ convolution in the residual connection aligns the channel dimensions.
The fused embedding $\boldsymbol{h}$ is then obtained through element-wise addition.
Built upon $\boldsymbol{h}$,
two parallel 1D convolution heads are deployed: 
an offset head predicts the offset $\boldsymbol{r}$, and a gate head outputs 
logits that are normalized by a sigmoid activation function to generate  the confidence gate $\boldsymbol{g} \in [0,1]$.
The final spatially refined gated offset is computed by element‑wise gating:
\begin{equation}
    \tilde{\Delta}\boldsymbol{X}^p_{sr} =
    \boldsymbol{g} \odot \boldsymbol{r},
\end{equation}
where $\odot$ denotes element‑wise multiplication.
Here, $\boldsymbol{g}$ controls the modulation strength at each sampling point, and $\boldsymbol{r}$ provides the corresponding correction.
This gated mechanism  selectively  amplifies corrections  on locally ambiguous segments while suppressing unnecessary perturbations in well-aligned regions.

Notably, 
the number of sampling points adopted by RoI Pooling for cross-layer refinement may be inconsistent with the lane-point sampling length $N$ utilized for lane representation.
To this end,
we employ 1D linear interpolation to resample the gated offset $\tilde{\Delta}\boldsymbol{X}^p_{sr}$ to the target length $N$,
yielding aligned spatially refined gated offset $\Delta \boldsymbol{X}^p_{sr}$.
Then the final horizontal coordinate update integrates anchor-based geometric constraints, fully connected regression residuals, and AGLR correction.
\begin{equation}
\boldsymbol{X}^p = \boldsymbol{X}^p + \Delta \boldsymbol{X}^p + \Delta \boldsymbol{X}^p_{sr}.
\label{eq:spatial_refine_update}
\end{equation}
This dual-path design decouples globally coarse raw geometry $\boldsymbol{X}^p + \Delta \boldsymbol{X}^p$ from local structure refinement $\Delta \boldsymbol{X}^p_{sr}$, which improves reliability to abrupt curvature changes and complex road topologies.

During inference, 
we further dynamically modulate the AGLR correction using the geometric fidelity $\hat{q}^*$, which is predicted at the final refinement stage and defined in Eq.\ref{Eq6}.
\begin{equation}
\Delta \boldsymbol{X}^p_{sr} \leftarrow
\left(1-\hat{q}^{*}\right)^{\gamma}\,\Delta \boldsymbol{X}^p_{sr},
\end{equation}
where $\gamma$ is a focal exponent. 
This strategy amplifies the correction for low geometric fidelity candidates and suppresses it for high geometric fidelity ones, leading to more stable and reliable lane geometry. 
The factor $(1-\hat{q}^{*})^{\gamma}$ acts as a reliability-conditioned policy.
When $\hat{q}^*$ is high, the lane prior already aligns well with the GT lane, so further adjustments are attenuated to avoid over-correction and local drift.
When $\hat{q}^*$ is low, the correction is preserved or even emphasized to improve insufficiently aligned candidates.
This spatial-level modulation mirrors the geometry-aware reweighting performed by the CRI in Eq.\ref{eq:fidelity_score_fusion} in the ranking process, thereby maintaining consistency between geometric refinement and candidate selection.

\subsection{Loss Function}
The overall training  objective integrates five loss terms:
\begin{equation}
\begin{aligned}
L_{all} &= w_{reg}L_{reg}+ w_{iou}L_{iou}  +w_{cls}L_{cls}+\\
&\quad w_{fid}L_{fid}+ w_{seg}L_{seg},
\end{aligned}
\end{equation}
where the scalar coefficients $w_{reg}, w_{iou}, w_{cls}, w_{fid}, w_{seg}$ balance  the contribution of each loss term.
The regression loss $L_{reg}$ penalizes the discrepancy between the predicted lane parameters $\hat{\boldsymbol{t}}_j$ of each positive lane prior and its matched GT lane $\boldsymbol{t}_j^{*}$, using the Smooth‑$L1$ loss.
The IoU loss $L_{iou}$ is computed  as $1 - \mathrm{LaneIoU}(\boldsymbol{X}_j^p, \boldsymbol{X}^g)$ for every positive lane prior, 
where $\boldsymbol{X}_j^{p}$ and $\boldsymbol{X}^g$ are the predicted and the matched GT lane horizontal coordinates, respectively.
The classification loss $L_{cls}$ optimizes the classification confidence $\hat{p}_j$.
The fidelity loss $L_{fid}$ aligns the predicted geometric fidelity, as described in Sec.\ref{lcc}.
The segmentation loss $L_{seg}$ is a per‑pixel cross‑entropy loss applied to the auxiliary segmentation head.
All detection losses are averaged over the mini‑batch and across all refinement stages.

\section{Experiments}

\subsection{Datasets}

We evaluate the proposed method on two widely recognized benchmark datasets, 
CULane\begingroup
\hypersetup{urlcolor=blue}%
\footnote{\url{https://xingangpan.github.io/projects/CULane}}%
\endgroup \cite{pan2018spatial} and 
CurveLanes\begingroup
\hypersetup{urlcolor=blue}%
\footnote{\url{https://github.com/SoulmateB/CurveLanes}}%
\endgroup \cite{CurveLane-NAS}.
Both  datasets encompass  diverse and challenging scenarios, 
which are essential for evaluating  the reliability and accuracy of lane detection.

CULane \cite{pan2018spatial} is widely regarded as one of the most challenging benchmarks.
It includes both normal driving conditions and eight challenging scenarios, 
such as nighttime driving and vehicle occlusions.
This diverse set of conditions enables comprehensive evaluation of lane detection methods across a wide range of real‑world environments. 
The dataset is collected using cameras mounted 
on six different vehicles, 
yielding over $55$ hours of footage from which $133$,$235$ frames are extracted.
To mitigate overfitting from highly similar consecutive frames, we adopt the de-duplication strategy from CLRerNet\cite{honda2024clrernet} on the training split, removing near-duplicates according to a pixel-difference threshold.
The resulting data comprise a training set of $55$,$698$ images, a validation set of $9$,$675$ images, and a testing set of $34$,$680$ images.

Compared to CULane,
CurveLanes \cite{CurveLane-NAS} focuses on scenes with high curvature and complex topologies, such as S-shaped curves, Y-shaped lanes, and multi-lane situations involving  more than eight lanes. 
These properties  establish a more demanding benchmark, particularly for evaluating intricate lane geometries in highly congested traffic.
The dataset is split into a training set of $100$,$000$ images, a validation set of $20$,$000$ images, and a testing set of $30$,$000$ images.
As the official testing set annotations have not been released, 
we report results on the validation set 
following \cite{liu2021condlanenet}.

\begin{table*}[!t]
    \centering
    \footnotesize 
    \setlength{\tabcolsep}{6pt}
    \renewcommand{\arraystretch}{0.92}
    \setlength{\cmidrulekern}{0pt}
    \caption{Quantitative Comparisons with SOTA Lane Detection Methods on the CULane Dataset. 
    The Reliability and Adaptability of the Proposed GFSR are Evaluated by Using Multiple Backbones across Nine Challenging Real-World Scenarios. False Positives are Reported for Cross Scenario, while \mbox{F1$_{50}$} Scores are Reported in Other Eight Scenarios.
    \textbf{Best Performance} and \underline{Second Best} are Highlighted.
    }
    \label{tab:culane_compact}
    \resizebox{1.02\textwidth}{!}{
    \begin{tabular}{c|c|cc|ccccccccc|cc}
    \toprule[0.7pt]
    \multirow{2}{*}{\raisebox{-0.85ex}{\textbf{Method}}} &
    \multirow{2}{*}{\raisebox{-0.85ex}{\textbf{Backbone}}} &
    \multicolumn{2}{c|}{\textbf{F1-Score(\%)}} &
    \multicolumn{9}{c|}{\textbf{Scenarios}} &
    \multicolumn{2}{c}{\textbf{Efficiency}} \\
    \cmidrule(l){3-4}\cmidrule{5-13}\cmidrule{14-15}
    & & \textbf{\textit{F1}}$_{\textbf{\textit{50}}}$ & \textbf{\textit{F1}}$_{\textbf{\textit{75}}}$ & \textbf{Normal} & \textbf{Crowd} & \textbf{Dazzle} & \textbf{Shadow} & \textbf{No line} & \textbf{Arrow} & \textbf{Curve} & \textbf{Cross}$\downarrow$ & \textbf{Night} & \textbf{\textit{GFLOPs}} & \textbf{\textit{FPS}} \\
    \midrule[0.4pt]
    SCNN\cite{pan2018spatial}                & VGG16       & $71.60$ & $39.84$ & $90.60$ & $69.70$ & $58.50$ & $66.90$ & $43.40$ & $84.10$ & $64.40$ & $1990$ & $66.10$ & $328.4$ & $39$   \\
    RESA\cite{zheng2021resa}                 & Res50       & $75.30$ & $53.39$ & $92.10$ & $73.10$ & $69.20$ & $72.80$ & $47.70$ & $88.30$ & $70.30$ & $1503$ & $69.90$ & $43.0$  & $-$  \\
    LaneATT\cite{Tabelini_2021_CVPR}         & Res122      & $77.02$ & $57.50$ & $91.74$ & $76.16$ & $69.47$ & $76.31$ & $50.46$ & $86.29$ & $64.05$ & $1264$ & $70.81$ & $70.5$  & $50$  \\
    UFLDv2\cite{qin2022ultra}                & Res34       & $76.00$ & $-$     & $92.50$ & $74.80$ & $65.50$ & $75.50$ & $49.20$ & $88.80$ & $70.10$ & $1910$ & $70.80$ & $-$     & $145$ \\
    CondLaneNet\cite{liu2021condlanenet}     & Res101      & $79.48$ & $61.23$ & $93.47$ & $77.44$ & $70.93$ & $80.91$ & $54.13$ & $90.16$ & $75.21$ & $1201$ & $74.80$ & $44.8$  & $79$ \\
    GANet\cite{wang2022keypoint}             & Res101      & $79.63$ & $-$     & $93.67$ & $78.66$ & $71.82$ & $78.32$ & $53.38$ & $89.86$ & $77.37$ & $1352$ & $73.85$ & $-$     & $-$  \\
    CLRKDNet\cite{qi2024clrkdnet}            & DLA34       & $80.68$ & $-$     & $93.86$ & $79.95$ & $75.63$ & $81.88$ & $54.85$ & $90.42$ & $72.81$ & $1147$ & $75.82$ & $-$     & $-$ \\
    CondLSTR\cite{Chen_2023_ICCV}            & Res101      & $80.77$ & $-$     & $94.17$ & $79.90$ & $75.43$ & $80.99$ & $55.00$ & $90.97$ & $76.87$ & $1047$ & $75.11$ & $50.2$  & $77$ \\
    LATR\cite{lv2024siamese}                 & Swin-B   & $80.85$ & $-$     & $93.92$ & $\underline{80.21}$ & $\underline{76.04}$ & $81.65$ & $55.42$ & $89.53$ & $75.66$ & $1043$ & $75.81$ & $16.8$  & $-$  \\
    DLNet\cite{lu2025dlnet}                  & DLA34       & $\underline{81.23}$ & $64.75$ & $94.10$ & $80.13$ & $\mathbf{76.24}$ & $\underline{83.97}$ & $\mathbf{56.77}$ & $90.69$ & $74.15$ & $1098$ & $76.18$ & $18.8$  & $104$  \\
    \midrule[0.4pt]
    \multirow{4}{*}{CLRNet\cite{zheng2022clrnet}}
                 & Res18       & $79.58$ & $62.21$ & $93.30$ & $78.33$ & $73.71$ & $79.66$ & $53.14$ & $90.25$ & $71.56$ & $1321$ & $75.11$ & $11.9$  & $112$  \\
                 & Res34       & $79.73$ & $62.11$ & $93.49$ & $78.06$ & $74.57$ & $79.92$ & $54.01$ & $90.59$ & $72.77$ & $1216$ & $75.02$ & $21.5$  & $87$  \\
                 & Res101      & $80.13$ & $62.96$ & $93.85$ & $78.78$ & $72.49$ & $82.33$ & $54.50$ & $89.79$ & $75.57$ & $1262$ & $75.51$ & $42.9$  & $57$   \\
                 & DLA34       & $80.47$ & $62.78$ & $93.73$ & $79.59$ & $75.30$ & $82.51$ & $54.58$ & $90.62$ & $74.13$ & $1155$ & $75.37$ & $18.4$  & $102$ \\
    \midrule[0.4pt]
    \multirow{5}{*}{CLRerNet\cite{honda2024clrernet}}
             & Res18       & $80.38$ & $63.70$ & $93.79$ & $78.69$ & $74.88$ & $82.59$ & $54.97$ & $90.38$ & $71.70$ & $1061$ & $75.81$ & $11.9$  & $129$   \\
             & Res34       & $80.76$ & $63.77$ & $93.93$ & $79.51$ & $73.88$ & $83.16$ & $55.55$ & $\underline{90.87}$ & $74.45$ & $1088$ & $\underline{76.53}$ & $21.5$  & $112$   \\
             & Res50       & $80.84$ & $64.06$ & $93.83$ & $79.74$ & $74.68$ & $83.27$ & $\underline{56.62}$ & $90.78$ & $73.55$ & $1256$ & $76.23$ & $23.9$  & $87$   \\
             & Res101      & $80.91$ & $64.30$ & $93.91$ & $80.03$ & $72.98$ & $82.92$ & $55.73$ & $90.53$ & $73.83$ & $1113$ & $76.13$ & $42.9$  & $57$  \\
             & DLA34       & $81.12$ & $64.07$ & $94.02$ & $80.20$ & $74.41$ & $83.71$ & $56.27$ & $90.39$ & $74.67$ & $1161$ & $\underline{76.53}$ & $18.4$  & $102$  \\
    \midrule[0.7pt]
    \multirow{6}{*}{\textbf{GFSR (Ours)}}
                         & Res18       & $80.64$ & $64.12$ & $93.91$ & $79.37$ & $72.03$ & $82.41$ & $53.56$ & $90.62$ & $75.59$ & $\underline{918}$  & $75.94$ & $12.4$  & $129$  \\
                         & Res34       & $80.90$ & $64.13$ & $94.09$ & $79.58$ & $73.63$ & $81.80$ & $55.47$ & $\mathbf{91.47}$ & $\underline{77.97}$ & $1045$ & $75.95$ & $21.7$  & $112$ \\
                         & Res50       & $81.14$ & $64.78$ & $94.06$ & $80.07$ & $74.60$ & $83.59$ & $55.44$ & $90.48$ & $76.57$ & $986$  & $76.02$ & $24.0$  & $87$  \\
                         & Res101      & $81.10$ & $\underline{64.80}$ & $93.97$ & $80.10$ & $72.59$ & $83.09$ & $55.74$ & $90.86$ & $77.64$ & $1045$ & $76.05$ & $43.9$  & $57$  \\
                         & RepViT      & $81.21$ & $64.49$ & $\mathbf{94.32}$ & $79.86$ & $73.96$ & $\mathbf{84.21}$ & $55.23$ & $90.75$ & $\mathbf{79.09}$ & $965$ & $76.17$ & $14.7$ & $117$ \\
                         & DLA34       & $\mathbf{81.46}$ & $\mathbf{65.01}$ & $\underline{94.27}$ & $\mathbf{80.22}$ & $75.99$ & $83.53$ & $55.39$ & $90.50$ & $77.71$ & $\mathbf{880}$ & $\mathbf{76.55}$ & $18.8$ & $102$ \\
    \bottomrule[0.7pt]
    \end{tabular}
    }
\end{table*}

\subsection{Evaluation Metrics}

Consistent with prior works \cite{behrendt2019unsupervised, zheng2022clrnet}, 
we adopt the F1 score as the primary  evaluation metric.
Specifically,
an IoU matrix is first calculated between predictions and GTs,
both represented as $30$-pixel-wide segmentation masks.
Linear sum assignment is then applied to establish one-to-one correspondences between predictions and GTs.
A matched prediction-GT pair is counted as a True Positive (TP) if its IoU exceeds a predefined threshold.
Otherwise, unmatched predictions are tallied as False Positives (FP) and False Negatives (FN).
The F1 score is subsequently  computed as
\begin{equation}
    F_1  =\frac{2*Prec.*Rec.}{Prec.+Rec.},
\end{equation}
where $Prec.=\frac{TP}{TP+FP}$ and $Rec.=\frac{TP}{TP+FN}$.
We report F1 score at two IoU thresholds ($0.5$ and $0.75$),
denoted as \mbox{F1$_{50}$} and \mbox{F1$_{75}$}, respectively.
The former captures  overall detection quality, while the latter penalizes imprecise lane boundaries, thereby providing insight into  localization accuracy.

In addition, 
we employ Giga Floating Point Operations Per second (GFLOPs)
to quantify the computational cost of inference. 
Given an input $Z$ and a network consisting of $L$ computational blocks, the complexity is defined as
\begin{equation}
GFLOPs(Z)=\frac{1}{10^9}\sum_{l=1}^{L}F_l(Z),
\end{equation}
where $F_l(Z)$ is the FLOPs of block $l$ for input $Z$.
This metric reflects the model's computational complexity and enables a fair comparison of resource requirements across different architectures. 
To further evaluate  practical inference speed,
we measure throughput in Frames Per Second (FPS),
i.e., the number of images processed per second.

\subsection{Implementation Details}

Our method is implemented in PyTorch with MMCV and trained on a single  NVIDIA RTX 4080 SUPER GPU with 32GB of memory. 
GFLOPs and FPS are measured on the same GPU under identical conditions.
All measurements use a batch size of $1$, a spatial resolution of $320 \times 800$, and CUDA‑synchronized timing after a warm‑up phase. 
The reported results are averaged over multiple repeated runs.

We employ ImageNet-pretrained ResNet \cite{he2016deep}, RepViT \cite{wang2024repvit}, and DLA \cite{8578353} as the backbone networks. 
The model is optimized using AdamW with an initial learning rate of $6 \times 10^{-4}$ and a cosine annealing schedule. 
For each dataset, training proceeds for $18$ epochs with a batch size of $24$. 
Our detector utilizes $192$ lane priors, each represented by $72$ points, and performs iterative refinement on $36$ sampled points.
For data augmentation during training, 
we apply a set of random transformations,
including random flipping, affine and scale‑rotation transforms, color jittering, and blur/compression perturbations.

We tune only a small set of dataset-specific hyperparameters. 
On CULane, the fidelity loss weight is set to $0.7$, 
and the fusion coefficients $(\beta_0, \beta_1)$ are set to $0.4$ and $0.6$, respectively. 
On CurveLanes, we increase the fidelity loss weight to $1.0$ and set $(\beta_0, \beta_1)$ to $0.6$ and $0.4$. 
Here, $(\beta_0, \beta_1)$ are calibration coefficients that fuse classification confidence with geometric fidelity to form the final collaborative reliability index. 
All remaining hyperparameters follow CLRerNet\cite{honda2024clrernet}.

\subsection{Quantitative Comparisons and Analysis}

We evaluate the proposed GFSR on the CULane and CurveLanes datasets and compare it rigorously against mainstream deep learning‑based lane detection methods.
The results are reported in Tab.\ref{tab:culane_compact} and Tab.\ref{tab:curvelanes_sota}, respectively.
To further assess reliability and adaptability, 
we test GFSR with a variety of backbones, including ResNet-18/34/50/101 \cite{he2016deep}, RepViT-M1.5 \cite{wang2024repvit}, and DLA34 \cite{8578353}.
The corresponding results are summarized in Tab.\ref{tab:culane_compact}.
In addition, a fine-grained analysis of high-curvature scenarios under multiple IoU thresholds is provided in Tab.\ref{tab:curve_fine_grained}, offering deeper insight into performance trends under stricter geometric matching criteria.

\subsubsection{Comparisons on CULane}
As illustrated in Tab.\ref{tab:culane_compact}, 
the proposed GFSR consistently improves performance under all evaluated backbone configurations on the CULane benchmark.
In contrast, 
CondLSTR \cite{Chen_2023_ICCV} and LATR \cite{lv2024siamese} still suffer from the mismatch between confidence prediction and geometric localization accuracy, as well as insufficient local refinement capability, in high-curvature lanes, complex scenario interference, and topology-changing regions,
which limits their performance under strict IoU thresholds.
Compared with CLRerNet\cite{honda2024clrernet}, 
our GFSR raises the \mbox{F1$_{50}$} and \mbox{F1$_{75}$} scores to $81.46\%$ and $65.01\%$,  respectively, 
corresponding to improvements of $0.34\%$ and $0.94\%$.
The larger gain on \mbox{F1$_{75}$} score demonstrates that 
our GFSR attains higher geometric fidelity under stricter overlap criteria.
Notably, our GFSR maintains real-time inference speed while incurring  only negligible additional computational overhead in terms of GFLOPs.

These gains originate from the complementary roles of its two core modules.
The LCC module explicitly predicts the geometric fidelity of lane candidates and uses it to recalibrate the confidence.
This recalibration makes the subsequent ranking and suppression more reliable, 
thereby preserving candidates that exhibit both high confidence and high geometric quality.
The AGLR module performs gated point‑wise adjustment along the predefined  rows, which improves fitting accuracy in geometrically challenging regions.

\begin{table*}[htbp]
    \centering
    \footnotesize 
    \setlength{\tabcolsep}{9pt}
    \renewcommand{\arraystretch}{0.92}
    \setlength{\cmidrulekern}{0pt}
    \caption{Quantitative Comparisons with SOTA Lane Detection Methods on the CurveLanes Dataset.
    \textbf{Best Performance} and \underline{Second Best} are Highlighted.}
    \label{tab:curvelanes_sota}
    \resizebox{0.63\textwidth}{!}{
    \begin{tabular}{ccccccc}
    \toprule[0.7pt]
    \textbf{Method} & \textbf{Backbone} & \textbf{\textit{F1}}$_{\textbf{\textit{50}}}$ & \textbf{\textit{Prec.}}$_{\textbf{\textit{50}}}$ & \textbf{\textit{Rec.}}$_{\textbf{\textit{50}}}$ & \textbf{\textit{GFLOPs}} & \textbf{\textit{FPS}} \\
    \midrule[0.4pt]
    SCNN\cite{pan2018spatial}                 & VGG16    & $65.02$ & $76.13$ & $56.74$ & $328.4$ & $-$   \\
    PointLaneNet\cite{chen2019pointlanenet}   & $-$        & $78.47$ & $86.33$ & $72.91$ & $14.8$  & $-$   \\
    CurveLane-L\cite{CurveLane-NAS}           & Searched & $82.29$ & $91.11$ & $75.03$ & $20.7$  & $-$   \\
    UFLDv2\cite{qin2022ultra}                 & Res34    & $81.34$ & $81.93$ & $80.76$ & $95.5$  & $145$  \\
    CondLaneNet\cite{liu2021condlanenet}      & Res101   & $86.10$ & $88.98$ & $\mathbf{83.41}$ & $44.9$  & $77$  \\
    CLRNet\cite{zheng2022clrnet}              & DLA34    & $86.10$ & $91.40$ & $81.39$ & $18.4$  & $101$  \\
    CANet-S\cite{yang2023canet}               & $-$        & $86.57$ & $91.37$ & $82.25$ & $13.1$  & $-$ \\
    CLRerNet\cite{honda2024clrernet}          & DLA34    & $86.47$ & $91.66$ & $81.83$ & $18.4$  & $101$  \\
    \midrule[0.7pt]
    \multirow{2}{*}{\textbf{GFSR (Ours)}}
                          & DLA34    & $\underline{87.03}$ & $\underline{91.67}$ & $82.83$ & $18.8$  & $100$ \\
                          & RepViT   & $\mathbf{87.35}$ & $\mathbf{92.13}$ & $\underline{83.05}$ & $14.7$  & $115$ \\
    \bottomrule[0.7pt]
    \end{tabular}
    }
\end{table*}
\begin{table*}[htbp]
    \centering
    \footnotesize
    \setlength{\tabcolsep}{7.7pt}
    \renewcommand{\arraystretch}{0.92}
    \setlength{\cmidrulekern}{0pt}
    \caption{Fine-Grained Comparisons in the Curve Scenario of the CULane Dataset. We Report the F1 Scores under IoU Thresholds Ranging from 0.50 to 0.90 (Step Size 0.05) to Evaluate Localization Reliability under Increasingly Strict Matching Criteria.}
    \label{tab:curve_fine_grained}
    \resizebox{0.79\textwidth}{!}{%
    \begin{tabular}{ccccccccccc}
    \toprule[0.7pt]
    \textbf{Method} & \textbf{Backbone} & \textbf{\textit{F1}}$_{\textbf{\textit{50}}}$ & \textbf{\textit{F1}}$_{\textbf{\textit{55}}}$ & \textbf{\textit{F1}}$_{\textbf{\textit{60}}}$ & \textbf{\textit{F1}}$_{\textbf{\textit{65}}}$ & \textbf{\textit{F1}}$_{\textbf{\textit{70}}}$ & \textbf{\textit{F1}}$_{\textbf{\textit{75}}}$ & \textbf{\textit{F1}}$_{\textbf{\textit{80}}}$ & \textbf{\textit{F1}}$_{\textbf{\textit{85}}}$ & \textbf{\textit{F1}}$_{\textbf{\textit{90}}}$ \\
    \midrule[0.4pt]
    CLRerNet\cite{honda2024clrernet} & DLA34 & $74.62$ & $70.53$ & $65.67$ & $58.43$ & $49.06$ & $37.14$ & $23.17$ & $10.48$ & $2.39$ \\
    \multirow{2}{*}{\textbf{GFSR (Ours)}}
     & DLA34 & $77.71$ & $74.89$ & $70.74$ & $65.67$ & $59.91$ & $50.48$ & $39.22$ & $22.68$ & $6.75$ \\
     & RepViT & $79.09$ & $75.32$ & $71.81$ & $66.75$ & $61.10$ & $53.90$ & $41.39$ & $24.85$ & $8.05$ \\
    \bottomrule[0.7pt]
    \end{tabular}
    }
\end{table*}

\subsubsection{Comparisons on CurveLanes}
Compared with CondLaneNet \cite{liu2021condlanenet} and CLRerNet \cite{honda2024clrernet}, 
Tab.\ref{tab:curvelanes_sota} reports that the proposed GFSR achieves superior overall performance on the more challenging CurveLanes benchmark.
Under the same DLA34 backbone,
our GFSR raises the \mbox{F1$_{50}$} score from $86.47\%$ to $87.03\%$ and recall from $81.83\%$ to $82.83\%$, indicating more effective preservation of valid lane candidates.
When equipped with RepViT‑M1.5, our GFSR further attains an \mbox{F1$_{50}$} score of $87.35\%$, a precision of $92.13\%$, and a recall of $83.05\%$, establishing the highest \mbox{F1$_{50}$} score among all compared methods.

The performance gains on CurveLanes are primarily driven by the improved recall. 
In scenarios with dense lane interactions and topological variations, 
the LCC module calibrates classification confidence to retain lane candidates with higher  geometric quality, thereby directly increasing recall.
The AGLR module further enhances the continuity of lane predictions over long distances in high‑curvature regions, yielding a more accurate fit to complex nonlinear lane geometries.

\subsubsection{Reliability Analysis across Multiple Scenarios}
Tab.\ref{tab:culane_compact} summarizes the substantial improvements achieved by our GFSR in various challenging real-world scenarios.  
For example,
in curve scenarios, the \mbox{F1$_{50}$} score increases from $74.67\%$ to $77.71\%$.
In crossing scenarios, the number of FP drops from $1161$ to $880$. 
In dazzle conditions, the \mbox{F1$_{50}$} score improves by $1.58\%$.

These gains indicate that our method not only improves overall metrics but also enhances reliability in complex driving conditions with high geometric complexity and strong local interference. 
The reasons are twofold.
First, the LCC module strengthens the dependence of ranking and suppression on geometric fidelity, thereby preserving higher-quality lane candidates and suppressing spurious high‑confidence detections.
Second, the AGLR module enhances  local refinement  for lanes with complicated  shapes, 
thereby achieving more precise boundary localization.

\subsubsection{Fine-Grained Analysis in High-Curvature Scenarios}
Tab.\ref{tab:curve_fine_grained} presents a fine‑grained analysis of geometric fitting quality by reporting F1 scores under IoU thresholds ranging from $0.5$ to $0.9$.
Compared with CLRerNet\cite{honda2024clrernet} under the same DLA34 backbone,
our GFSR improves the F1 score from $74.62\%$ to $77.71\%$ at an IoU of $0.5$, 
from $37.14\%$ to $50.48\%$ at $0.75$, 
and from $2.39\%$ to $6.75\%$ at $0.9$. 
A similar pattern is observed with the RepViT‑M1.5 backbone.

Notably, the improvement becomes more pronounced as the IoU threshold increases, indicating gains in both presence detection and geometric fidelity. 
This is achieved by our two core designs: the LCC module retains candidates with higher geometric quality during confidence ranking, while the AGLR module enhances point‑wise fitting continuity in high‑curvature regions.

\subsubsection{Backbone Sensitivity Analysis}
As indicated in Tab.\ref{tab:culane_compact}, 
the proposed GFSR consistently surpasses the compared methods under various backbone configurations, including ResNet-18/34/50/101\cite{he2016deep}, DLA34\cite{8578353}, and RepViT-M1.5\cite{wang2024repvit}. 
Notably, when equipped with RepViT-M1.5,
our GFSR achieves the highest overall performance across multiple scenarios, attaining an \mbox{F1$_{50}$} of $79.09\%$ in the curve scenario, $94.32\%$ for normal scene, and $84.21\%$ for shadow scene. 
For applications prioritizing stricter matching criteria and cross‑scene reliability,
DLA34 emerges as the most effective backbone, 
delivering  an \mbox{F1$_{50}$} score of $65.01\%$ and reducing FP in cross scene to $880$. 
These results demonstrate that our method is not limited to a specific architecture but can be effectively deployed on both classic convolutional neural network and lightweight mobile networks.

\subsection{Qualitative Comparisons and Analysis}
\begin{figure}[t]
    \centering
    \includegraphics[width=\linewidth]{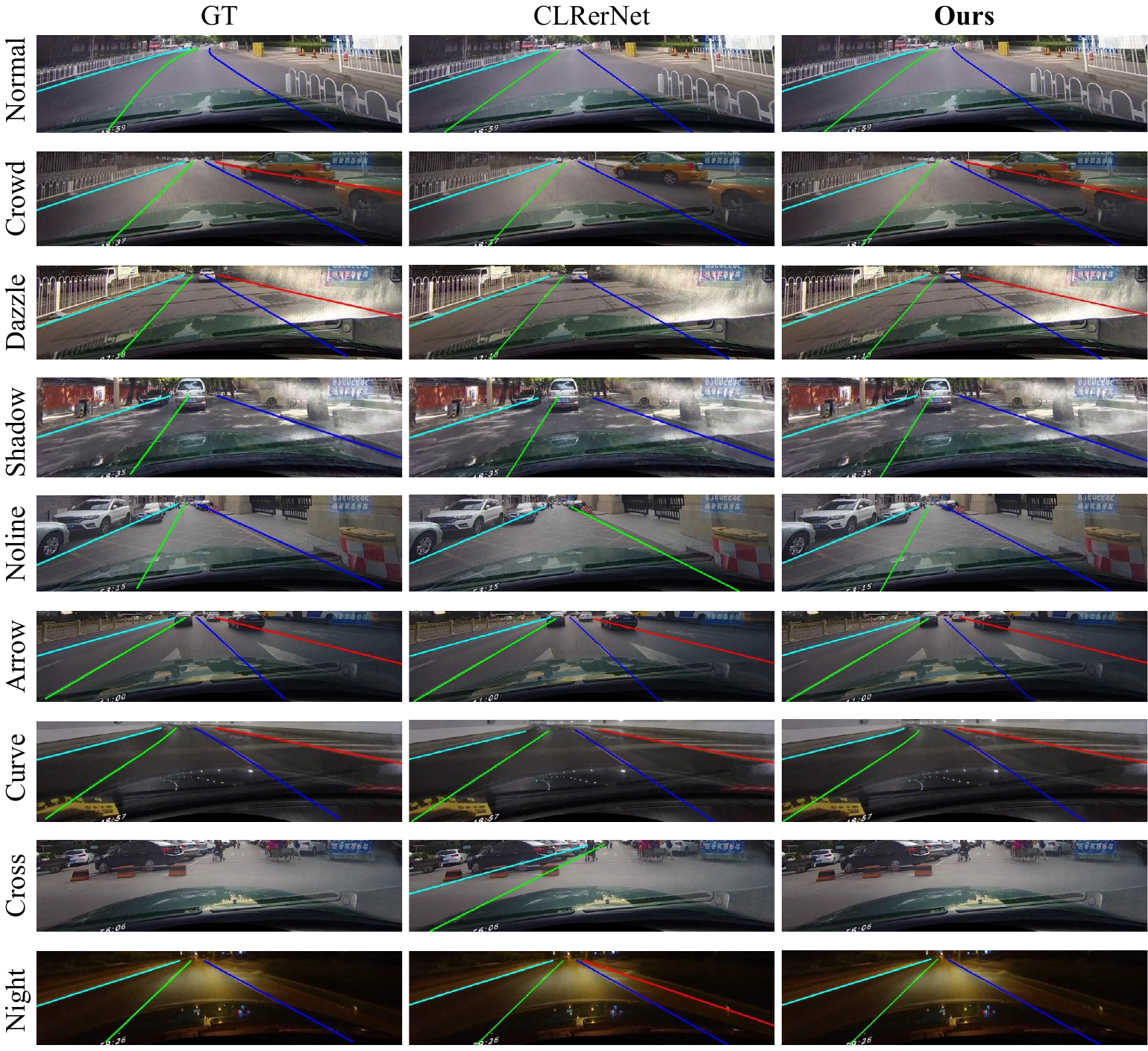}
    \caption{Qualitative comparison of detection results between our method and CLRerNet \cite{honda2024clrernet} under the same backbone (DLA34) across nine challenging scenarios. 
    All examples are taken from the testing set of the CULane benchmark.}
    \label{fig:fig5}
\end{figure}

To provide a more intuitive comparison, 
we visualize the lane detection results of CLRerNet \cite{honda2024clrernet} and our GFSR on the CULane and CurveLanes datasets, using DLA34 \cite{8578353} as the backbone.
The results are presented in Fig.\ref{fig:fig5} and Fig.\ref{fig:fig6}, respectively.
On CULane, 
we select nine challenging scenarios to evaluate geometric continuity, reliability to strong disturbances, and error recovery. 
On CurveLanes, we examine diverse lane shapes, including numerous curves and complex multi‑lane interactions, to assess whether the detected lane lines better preserve structural integrity.

\subsubsection{Comparisons on CULane}
Compared with CLRerNet \cite{honda2024clrernet},
Fig.\ref{fig:fig5} illustrates that lanes predicted by our GFSR are smoother and exhibit stronger geometric continuity, especially in regions with high curvature, strong visual interference, or dramatic topological changes.
CLRerNet \cite{honda2024clrernet} selects lane candidates primarily based on classification confidence, without explicitly accounting for geometric quality.
Moreover, its regression process lacks inter-point geometric constraints and point-wise refinement mechanisms,
making it difficult to accurately fit highly curved lane lines. 
As a result, CLRerNet \cite{honda2024clrernet} often produces nearly straight approximations instead of faithfully following the true curved shape, particularly near distant regions with abrupt curvature variations. 
In contrast, our GFSR more reliably  retains lane candidates with higher geometric fidelity, leading to substantially improved structural integrity.

In challenging scenarios with strong interference and significant topological changes, 
such as highlight, arrow, noline, night and crossing, 
CLRerNet \cite{honda2024clrernet} is vulnerable to noise and struggles to model complex topological structures. 
In contrast, our GFSR effectively improves lane fitting quality and reduces FP and FN.
\begingroup
\hypersetup{urlcolor=blue}%
To better showcase the qualitative behavior of our method, we provide video-based visualization results of GFSR on CULane in this repository\footnote{\url{https://github.com/71yvonne/gfsr-2d-lane-detection}}.%
\endgroup

\subsubsection{Comparisons on CurveLanes}

\begin{figure}[htbp]
    \centering
    \includegraphics[width=\linewidth]{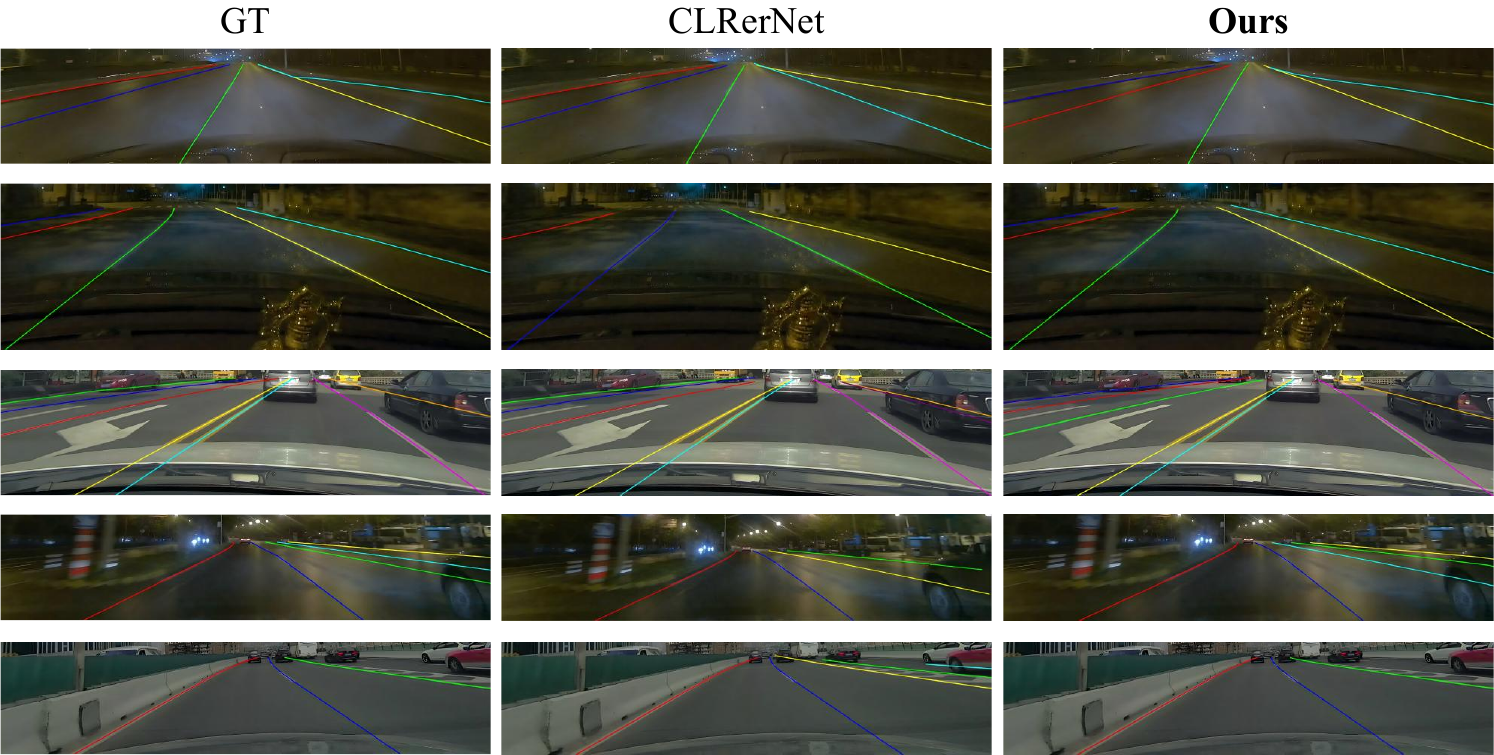}
    \caption{Qualitative comparison of detection results between our method and CLRerNet \cite{honda2024clrernet} under the same backbone (DLA34). All examples are taken from the validation set of the CurveLanes benchmark.}
    \label{fig:fig6}
\end{figure}

As shown in Fig.\ref{fig:fig6},
consistent with the trends observed on CULane, 
CLRerNet \cite{honda2024clrernet}
struggles to fully fit lane lines under high‑curvature and multi‑lane conditions on CurveLanes.
It is more susceptible  to FP and FN when facing more drastic curvature changes and denser lanes.
In contrast, the detection results predicted by our GFSR align more closely with the GTs through confidence calibration and point-wise refinement.


\begin{table}[htbp]
\centering
\footnotesize
\begin{minipage}{0.42\textwidth}
\centering
\setlength{\tabcolsep}{10.7pt}
\caption{Ablation Study of the Proposed Modules on CULane.}
\label{tab:ablation_culane}
\begin{threeparttable}
\resizebox{\linewidth}{!}{%
\begin{tabular}{ccccc}
\toprule[0.7pt]
 \textbf{LCC} & \textbf{AGLR} & \textbf{\textit{F1}}$_{\textbf{\textit{50}}}$ & \textbf{\textit{F1}}$_{\textbf{\textit{75}}}$ & \textbf{\textit{F1}}$_{\textbf{\textit{90}}}$ \\
\midrule[0.4pt]
  &  & $81.08$ & $64.12$ & $23.88$ \\
  \checkmark & & $81.29$ & $65.00$ & $\mathbf{24.88}$ \\
   & \checkmark & $81.18$ & $64.37$ & $24.03$ \\
 \checkmark & \checkmark & $\mathbf{81.46}$ & $\mathbf{65.01}$ & $24.67$ \\
\bottomrule[0.7pt]
\end{tabular}%
}
\begin{tablenotes}[flushleft]
\item \textbf{Note}: All improvements in Tab.\ref{tab:ablation_culane} are statistically significant (paired t-test, $p < 0.05$).
\end{tablenotes}
\end{threeparttable}
\end{minipage}
\end{table}
\subsubsection{Analysis of Module Contributions}
\begin{figure}[htbp]
    \centering
    \includegraphics[width=\linewidth]{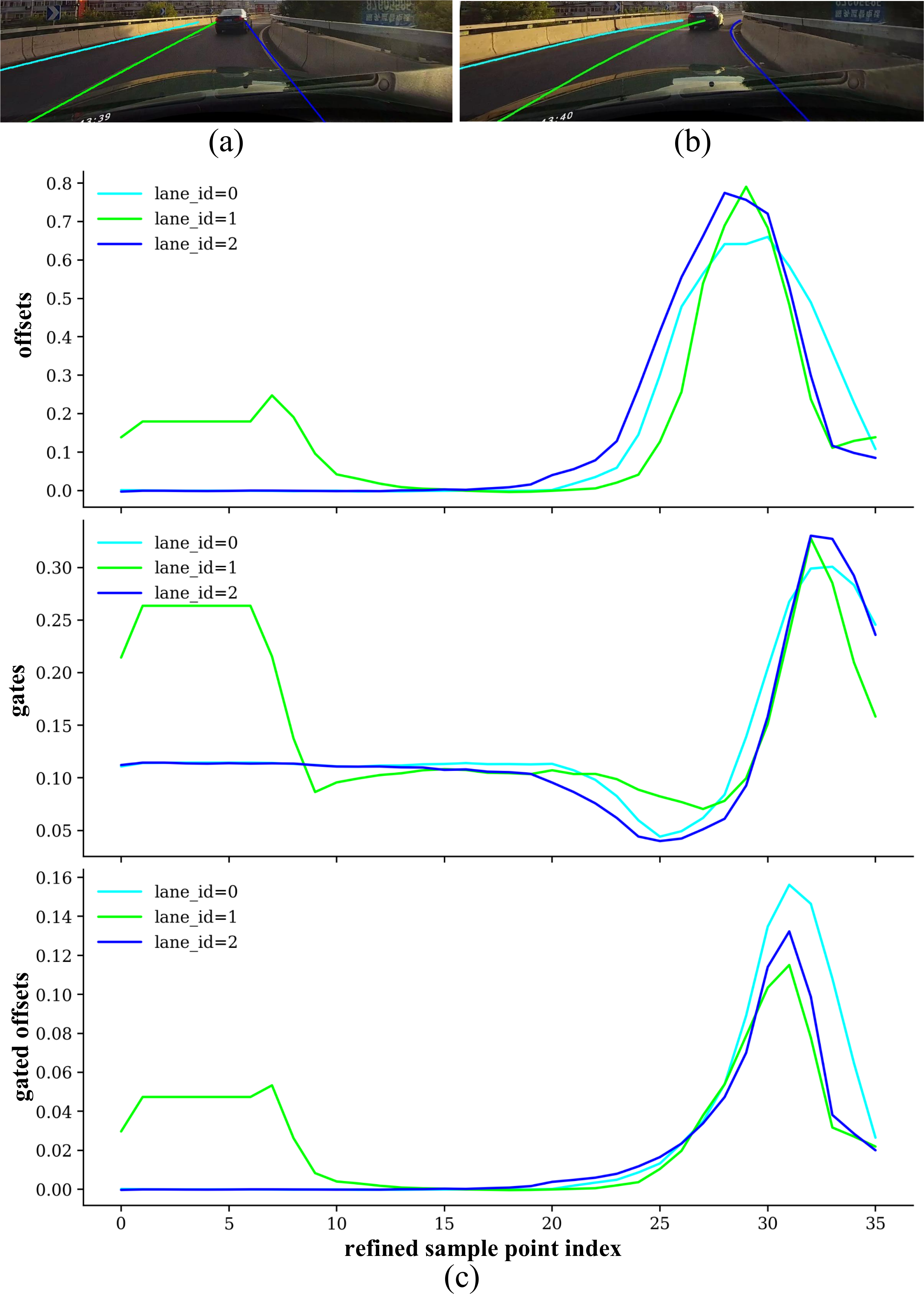}
    \caption{Qualitative visualization of the AGLR module in a curve-dominant scenario. 
    (\textbf{a}): Detection result from CLRerNet\cite{honda2024clrernet}.
    (\textbf{b}): Detection result after performing GFSR; 
    (\textbf{c}): Per-sample predictions of lane instances output by the AGLR module include offsets, confidence gates after sigmoid activation, and the final gated offsets obtained via element-wise multiplication.
    The visualization demonstrates that our AGLR module applies stronger corrections in challenging segments while preserving reliability in well-aligned regions.}
    \label{fig:fig7}
\end{figure}

We analyze how the proposed LCC and AGLR modules jointly improve performance in high‑curvature scenarios.
As shown in Fig.\ref{fig:fig7}(c), 
the offset branch predicts offsets along the lane lines. 
The gating mechanism is predominantly activated in geometrically challenging segments, 
such as those with high curvature, partial occlusion, or local blur,
generating amplified gated offsets while suppressing unnecessary adjustments elsewhere.
This adaptive point-wise refinement yields visibly smoother and more stable lane geometry, as illustrated in Fig.\ref{fig:fig7}(b), which helps explain the substantial gains in curve-dominated scenes.

Complementary to this geometric correction, the LCC module calibrates the classification confidence by fusing it with geometric fidelity, as evidenced by the correlation analysis in Fig.\ref{fig:fig1}(d). 
The CRI exhibits a markedly stronger correlation with LaneIoU than the original classification confidence alone. 
Consequently, subsequent operations such as filtering, ranking, and NMS become better aligned with CRI, establishing a reliable foundation for the final detection performance.

\subsection{Ablation Study}

\subsubsection{Effectiveness of the Proposed Module}
We conduct ablation studies on CULane and CurveLanes
to verify the effectiveness of the proposed LCC and AGLR modules.
The results are reported in Tab.\ref{tab:ablation_culane} and Tab.\ref{tab:ablation_curvelanes}, respectively. 
On CULane,
the LCC module mainly enhances performance under strict IoU thresholds,
raising the \mbox{F1$_{75}$} score from $64.12\%$ to $65.00\%$.
Combining both modules yields the best overall performance, with an \mbox{F1$_{50}$} score of $81.46\%$ and an \mbox{F1$_{75}$} score of $65.01\%$. 
The similar complementary effect emerges in Tab.\ref{tab:ablation_curvelanes}.
The LCC module increases precision from $91.18\%$ to $91.78\%$, whereas the AGLR module raises recall from $81.96\%$ to $82.47\%$.
Combining both modules achieves the highest \mbox{F1$_{50}$} score of $87.03\%$.

\subsubsection{Fidelity Loss}

Tab.\ref{tab:ablation_loss_curvelanes} presents an ablation study that investigates BCE loss and Smooth-$L_1$ loss as alternative choices for the fidelity objective $L_{fid}$ in Eq.\ref{eq:fidelity_loss}, and evaluates their performance under different loss weight configurations.
Both losses achieve optimal performance at a weight of $1.0$, 
with BCE slightly surpassing Smooth-$L_1$ at its best configuration.
This result indicates that BCE is more effective at aligning the predicted geometric fidelity with the soft LaneIoU supervision target in our implementation.
Consequently, confidence calibration is stabilized during training, 
leading to more reliable ranking behavior at inference.

\subsubsection{Multi-Stage vs. Single-Stage Insertion of the AGLR Module}

Tab.\ref{tab:ablation_spatial_culane} investigates the placement of the AGLR module within our framework.
A single-stage insertion improves particular metrics, 
such as stronger \mbox{F1$_{75}$} score at the $2^{nd}$ stage or stronger \mbox{F1$_{90}$} score at the $1^{st}$ stage.
In contrast, deploying the module across all stages achieves  the best overall balance,
delivering  the highest \mbox{F1$_{50}$} score of $81.46\%$ and consistently strong performance at strict IoU thresholds.
These results confirm that 
multi-stage refinement mitigates accumulated geometric errors more effectively than localized, single-stage optimization.


\begin{table}[htbp]
\centering
\footnotesize
\begin{minipage}{0.44\textwidth}
\centering
\setlength{\tabcolsep}{10.7pt}
\caption{Ablation Study of the Proposed Modules on CurveLanes.}
\label{tab:ablation_curvelanes}
\begin{threeparttable}
\resizebox{\linewidth}{!}{%
\begin{tabular}{ccccc}
\toprule[0.7pt]
 \textbf{LCC} & \textbf{AGLR} & \textbf{\textit{F1}}$_{\textbf{\textit{50}}}$ & \textbf{\textit{Prec.}}$_{\textbf{\textit{50}}}$ & \textbf{\textit{Rec.}}$_{\textbf{\textit{50}}}$ \\
\midrule[0.4pt]
  &  & $86.32$ & $91.18$ & $81.96$ \\
  \checkmark & & $86.39$ & $\mathbf{91.78}$ & $81.59$ \\
   & \checkmark & $86.78$ & $91.57$ & $82.47$ \\
 \checkmark & \checkmark & $\mathbf{87.03}$ & $91.67$ & $\mathbf{82.83}$ \\
\bottomrule[0.7pt]
\end{tabular}%
}
\begin{tablenotes}[flushleft]
\item \textbf{Note}: All improvements in Tab.\ref{tab:ablation_curvelanes} are statistically significant (paired t-test, $p < 0.05$).
\end{tablenotes}
\end{threeparttable}
\end{minipage}
\end{table}

\begin{table}[htbp]
\centering
\footnotesize
\setlength{\tabcolsep}{9.7pt}
\caption{Ablation Study of Fidelity Loss on CurveLanes.}
\label{tab:ablation_loss_curvelanes}
\resizebox{0.47\textwidth}{!}{%
\begin{tabular}{ccccc}
\toprule[0.7pt]
\textbf{Loss} & \textbf{Weight} & \textbf{\textit{F1}}$_\textbf{\textit{50}}$ & \textbf{\textit{Prec.}}$_\textbf{\textit{50}}$ & \textbf{\textit{Rec.}}$_\textbf{\textit{50}}$ \\
\midrule[0.4pt]
 & $0.7$ & $86.85$ & $91.73$ & $82.47$ \\
 & $1.0$ & $87.01$ & $91.87$ & $82.64$ \\
\textit{smooth-$l_1$} & $1.5$ & $87.00$ & $\mathbf{91.96}$ & $82.54$ \\
 & $2.0$ & $86.88$ & $91.75$ & $82.51$ \\
 & $2.5$ & $86.89$ & $91.75$ & $82.52$ \\
\midrule[0.4pt]
 & $0.3$ & $86.90$ & $91.89$ & $82.41$ \\
 & $0.5$ & $86.94$ & $91.88$ & $82.50$ \\
\textit{BCE} & $0.7$ & $86.83$ & $91.76$ & $82.39$ \\
 & $1.0$ & $\mathbf{87.03}$ & $91.67$ & $\mathbf{82.83}$ \\
 & $1.2$ & $86.95$ & $91.93$ & $82.48$ \\
\bottomrule[0.7pt]
\end{tabular}
}
\end{table}
\begin{table}[!t]
\centering
\footnotesize
\setlength{\tabcolsep}{8.7pt}
\caption{Ablation Study of AGLR Stage Placement on CULane.}
\label{tab:ablation_spatial_culane}
\resizebox{0.47\textwidth}{!}{%
\begin{tabular}{cccccc}
\toprule[0.7pt]
\textbf{Stage$_\textbf{0}$} & \textbf{Stage$_\textbf{1}$} & \textbf{Stage$_\textbf{2}$} & \textbf{\textit{F1}}$_\textbf{\textit{50}}$ & \textbf{\textit{F1}}$_\textbf{\textit{75}}$ & \textbf{\textit{F1}}$_\textbf{\textit{90}}$ \\
\midrule[0.4pt]
\checkmark &        &        & $81.24$ & $64.81$ & $24.70$ \\
           & \checkmark &        & $80.85$ & $65.09$ & $\mathbf{24.97}$ \\
           &        & \checkmark & $81.33$ & $\mathbf{65.34}$ & $23.69$ \\
\checkmark &        & \checkmark & $81.20$ & $64.69$ & $21.57$ \\
           & \checkmark & \checkmark & $81.40$ & $64.77$ & $21.63$ \\
\checkmark & \checkmark & \checkmark & $\mathbf{81.46}$ & $65.01$ & $24.67$ \\
\bottomrule[0.7pt]
\end{tabular}
}
\end{table}

\section{Conclusion}
We propose geometric fidelity and spatial refinement, a lightweight lane detection framework that couples LaneIoU-guided confidence calibration with adaptive gated location refinement. 
The former module estimates  geometric fidelity under the supervision of LaneIoU, and integrates it with classification confidence to construct the collaborative reliability index. With this index, threshold filtering and NMS ranking become aligned with geometric quality, rather than relying solely on classification confidence.
The latter module works in conjunction with the regression head during  each refinement stage to enhance  point-wise coherence and improve fitting in challenging lane regions. 
Extensive experiments on CULane and CurveLanes validate the effectiveness of the proposed framework.

\bibliographystyle{IEEEtran}

\bibliography{refs}

\end{document}